\documentclass{ieeeaccess}
\usepackage{cite}
\usepackage{amsmath,amssymb,amsfonts}
\usepackage{algorithmic}
\usepackage{graphicx}
\usepackage{textcomp}
\usepackage{subfig}
\usepackage{multirow}
\usepackage{booktabs}
\def\BibTeX{{\rm B\kern-.05em{\sc i\kern-.025em b}\kern-.08em
    T\kern-.1667em\lower.7ex\hbox{E}\kern-.125emX}}
\begin{document}
\doi{}

\title{Probabilistic Forecasting of Sensory Data with Generative Adversarial Networks -- ForGAN}
\author{\uppercase{Alireza Koochali\authorrefmark{1,2},
Peter Schichtel\authorrefmark{2}, Sheraz Ahmed\authorrefmark{3}, and Andreas Dengel\authorrefmark{1,3}}.}
\address[1]{Department of Computer Science, University of Kaiserslautern, 67663 Germany}
\address[2]{Ingenieurgesellschaft Auto und Verkehr (IAV), Kaiserslautern, 67663 Germany}
\address[3]{German Research Center for Artificial Intelligence (DFKI), Kaiserslautern, 67663 Germany}


\markboth
{Koochali \headeretal: Probabilistic Forecasting of Sensory Data with Generative Adversarial Networks -- ForGAN}
{Koochali \headeretal: Probabilistic Forecasting of Sensory Data with Generative Adversarial Networks -- ForGAN}

\corresp{Corresponding author: Alireza Koochali (e-mail: akoochal@ rhrk.uni-kl.de).}

\begin{abstract}
Time series forecasting is one of the challenging problems for humankind. Traditional forecasting methods using mean regression models have severe shortcomings in reflecting real-world fluctuations. While new probabilistic methods rush to rescue, they fight with technical difficulties like quantile crossing or selecting a prior distribution. To meld the different strengths of these fields while avoiding their weaknesses as well as to push the boundary of the state-of-the-art, we introduce ForGAN – one step ahead probabilistic forecasting with generative adversarial networks. ForGAN utilizes the power of the conditional generative adversarial network to learn the data generating distribution and compute probabilistic forecasts from it. We argue how to evaluate ForGAN in opposition to regression methods. To investigate probabilistic forecasting of ForGAN, we create a new dataset and demonstrate our method abilities on it. This dataset will be made publicly available for comparison. Furthermore, we test ForGAN on two publicly available datasets, namely Mackey-Glass dataset~\cite{mackey1977oscillation} and Internet traffic dataset (A5M)~\cite{cortez2012multi} where the impressive performance of ForGAN demonstrate its high capability in forecasting future values.
\end{abstract}


\titlepgskip=-15pt

\maketitle

\section{Introduction}
\label{sec:introduction}
At its core, life is about decision making. Decision making always depends on our perspective of the future. Therefore, the forecast of what might lay before us is one of the most intriguing challenges for humankind. It is no surprise that there is a huge and diverse community concerned with forecasting and decision making. To name, but a few, there is weather and climate prediction~\cite{racah2017extremeweather,rodrigues2018deepdownscale}, flood risk assessment~\cite{wiesel2018ml}, seismic hazard prediction~\cite{DBLP:journals/corr/abs-1810-01965,DBLP:journals/corr/abs-1809-02880}, predictions about the availability of (renewable) energy resources~\cite{gensler2018multi,DBLP:journals/corr/ChenWKZ17}, economic and financial risk management~\cite{DBLP:journals/corr/abs-1811-10791,DBLP:journals/corr/abs-1811-03711}, health care~\cite{DBLP:journals/corr/abs-1812-00371,DBLP:journals/corr/abs-1811-12234,DBLP:journals/corr/abs-1711-06402}, predictive and preventative medicine~\cite{zitnik2019machine} and many more. 

Since the forecast is the prediction of future values, we can take the predictive view of regression to provide a solution to this problem~\cite{gneiting2014probabilistic}. The ultimate goal of regression analysis is to obtain information about the conditional distribution of a response given a set of explanatory variables~\cite{hothorn2014conditional}. In the context of forecasting, the ultimate goal is obtaining the predictive probability distribution over future quantities or events of interest. In other words, given the time-dependent observable of interest $x$, the goal is to acquire $\rho(x_{t+1}|\{x_t,\ldots,x_0\})$. Unfortunately, the ultimate goal is seldom achieved and most of the methods focus on only one designated quantity of the response distribution, namely the mean. The mean regression models have the advantage of being easy to understand and predict however they often lead to incomplete analyses when more complex relationships are presented and also bears the risk of false conclusions about the significance/importance of covariates~\cite{kneib2013beyond}. In the following sections, we review mean regression forecast methods briefly and then we provide an overview about scientific endeavors on probabilistic forecasting. Note that in this paper, we call the history of events $\{x_{t},\ldots,x_0\}$ the condition $c$ and we use $x_{t+1}$ as the notion for the value of the next step i.e. target value.

\subsection{Mean Regression Forecast}
\begin{figure*}[t]
	\centering
	\subfloat[A cluster of time windows which are almost similar on every time step except for the value at the last step $x_{10}$.\label{fig:example_1}]{%
		\includegraphics[width=0.45\linewidth]{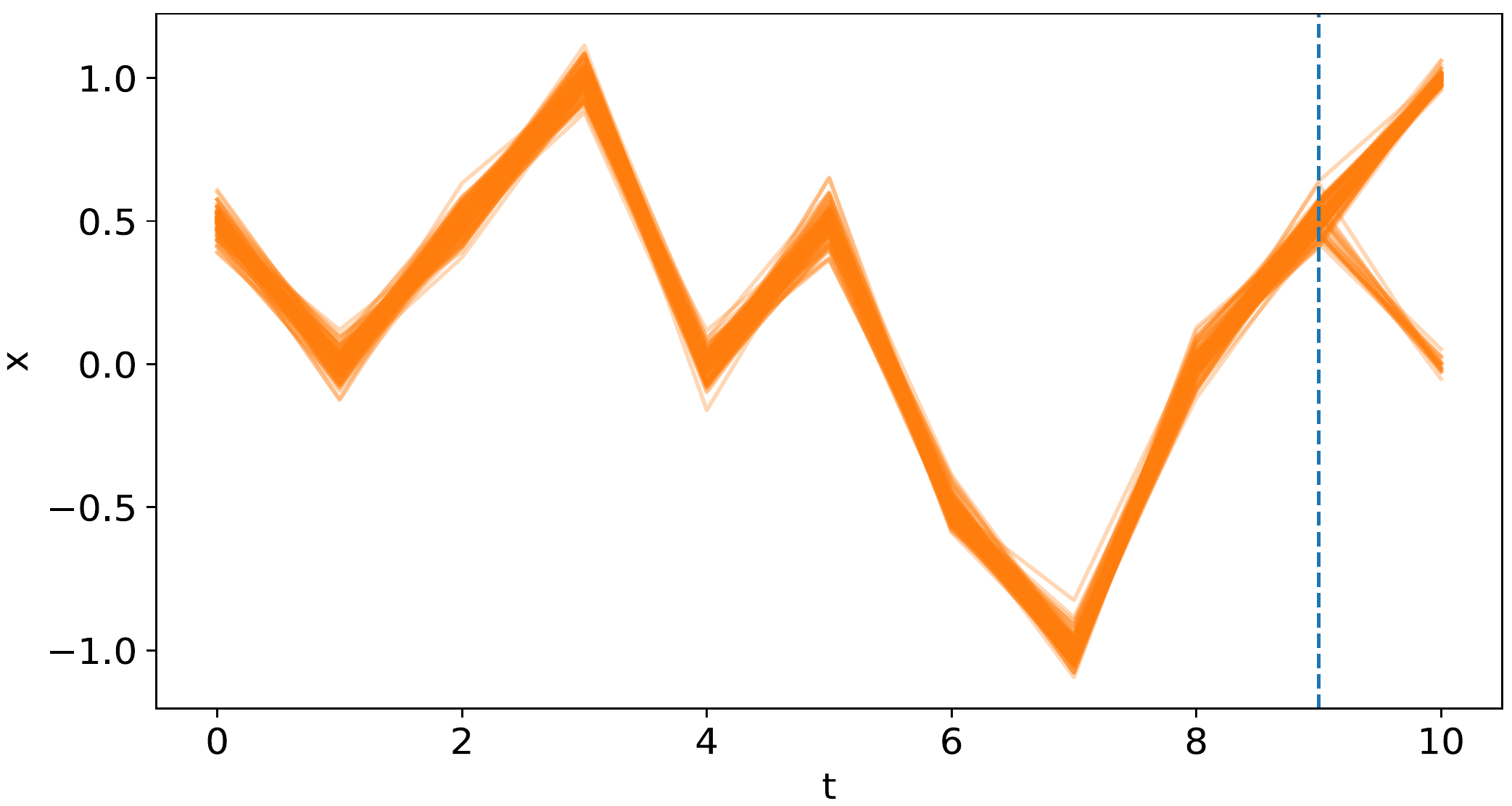}}
	\hfill
	\subfloat[The probability distribution of $x_{10}$ (in orange color) alongside the distribution learned by a mean regression model (in blue color).\label{fig:example_2}]{%
		\includegraphics[width=0.45\linewidth]{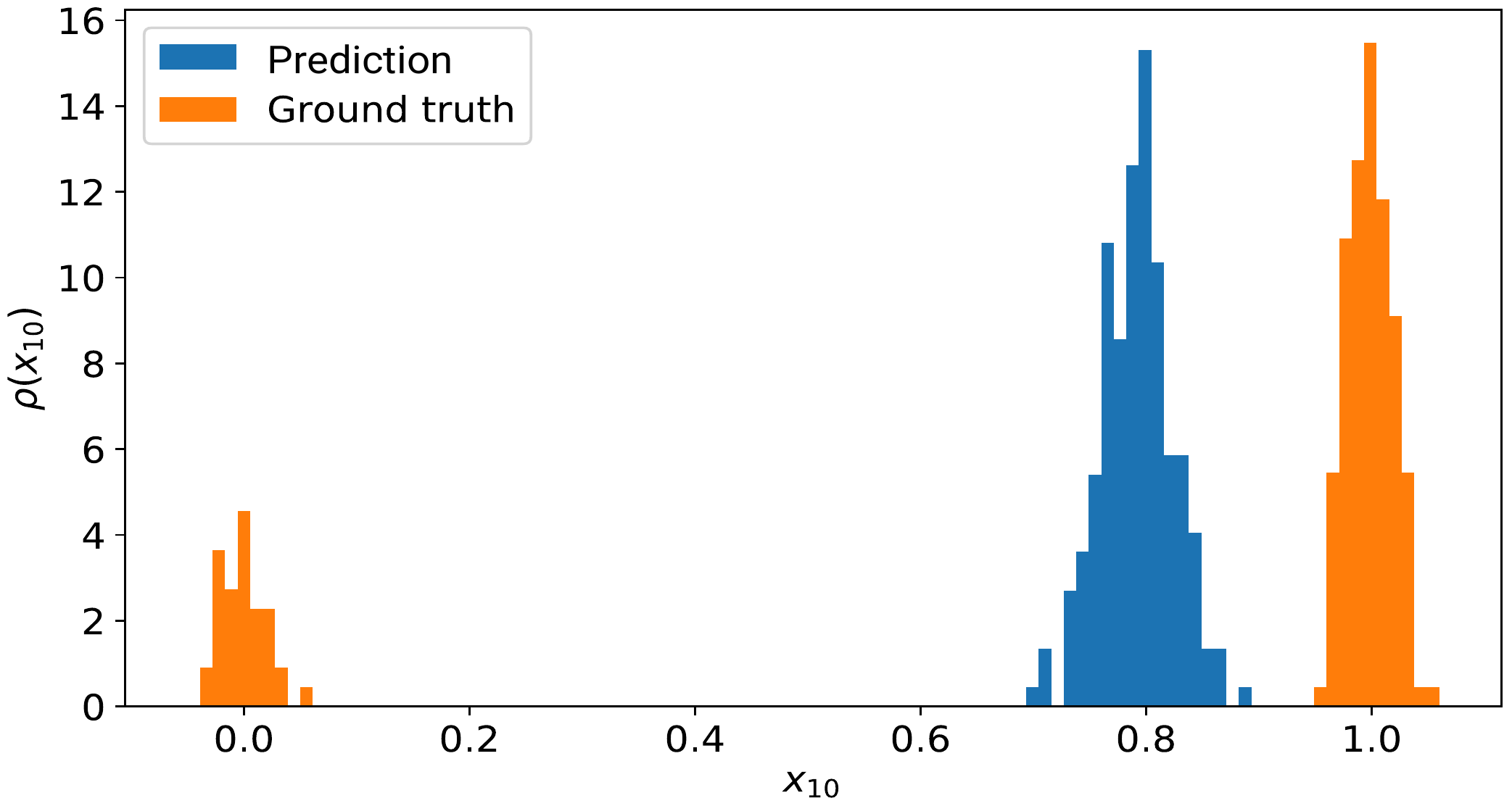}}
	\caption{(a) Dataset used for training mean regression and (b) visualization of the results.}
	\label{fig:example} 
\end{figure*}

Mean regression forecasting is concerned with predicting $\mu(\rho(x_{t+1}|c))$ most accurately. There is a broad range of mean regression methods available in literature e.g., statistical methods (like ARMA and ARIMA~\cite{Box1990TSA574978} and their variants), machine learning based methods (like Support Vector Machines (SVM)~\cite{yan2013comparison,yan2014mid,rubio2011heuristic,vapnik1998statistical,frohlich2003feature,huang2006ga}, Evolutionary Algorithms (EA)~\cite{cortez2001genetic,frohlich2003feature,huang2006ga,gan2012hybrid,kim2000genetic,cai2013novel,bas2014modified} and Fuzzy Logic Systems (FLS)~\cite{chu1996application,song1995new,egrioglu2013fuzzy,cai2013novel,bas2014modified,shah2012fuzzy,aladag2012new}), and Artificial Neural Network based  methods (ANN)~\cite{assaad2008new,ogunmolu2016nonlinear,dorffner1996neural,malhotra2015long}. These methods use handcrafted features on the data except ANNs which try to automatically extract those features using an end-to-end pipeline. As these methods forecast the future following the principles of mean regression, all of them inherit the main problem/limitation of these principles, i.e. they do not include the fluctuations around the mean value. Hence, their results can be unreliable and misleading in some cases. Fig.~\ref{fig:example_1} presents an example of the problem inherent in all mean regression based methods. It shows a cluster of time series with identical, but noisy, time window $c=\{x_9,\ldots,x_0\}$ and the future value at $t = 10$ (to be found right of the blue dashed line) which can take two distinctive realizations: in $80\%$ of the cases $x_{10}$ yields one while in $20\%$ of the cases it yields zero\footnote{If you like you may imagine this as an experiment on a chaotic system.}.

To demonstrate the problem, we train a simple neural network to forecast $x_{10}$. We present the result in Fig.~\ref{fig:example_2}. It illustrates that the regression model fails to model the data. The best answer we can get from mean regression will converge to $0.8$, the weighted average of all possible values for $x_{10}$. We can observe from Fig.~\ref{fig:example_2} that the values forecasted with mean regression do not have any overlap with ground truth. It indicates mean regression is incapable of predicting any ground truth value precisely and it cannot be improved any further. Note, this does not imply that the mean regression method does not work. The closer the target distribution approximates a Dirac delta distribution, the better and more accurate a mean regression forecast will be. It is upon the researcher to evaluate this constraint carefully.

\subsection{Probabilistic Forecast}

To solve the shortcomings associated with mean regression, recently many researchers presented solutions which are moving from mean regression to probabilistic forecasting. Probabilistic forecasting serves to quantify the variance in a prediction~\cite{gneiting2014probabilistic}. Different approaches have been proposed to undertake probabilistic forecasting in various fields~\cite{collins2007ensembles, gneiting2005weather, palmer2002economic, palmer2012towards, cloke2009ensemble, krzysztofowicz2001case, jordan2011operational, pinson2013wind, zhu2012short, groen2013real, timmermann2000density, montgomery2012ensemble, alkema2007probabilistic, raftery2012bayesian, jones2012improved, hood2004systems}. In the twentieth century, Stigler~\cite{stigler1975transition} coined the idea of the transition from point estimation to distribution estimation. However, the shift toward applying probabilistic forecasting on real-world problems did not take place until recent years. Two of the most prominent approaches in these fields are conditional quantile regression and conditional expectile regression. Quantile regression is a statistical technique intended to estimate, and conduct inference about, conditional quantile functions~\cite{koenkerquantile}. To estimate the regression coefficients from training data, one uses the asymmetric piecewise linear scoring function, which is consistent for the $\alpha$-quantile~\cite{koenker1978regression,koenkerquantile}. Expectile regression works similarly but it is based on the asymmetric piecewise quadratic scoring function~\cite{efron1991regression,newey1987asymmetric,sobotka2012geoadditive}. While these methods push regression methods beyond the mean regression, the problem of crossing quantile curves is frequently observed especially when considering a dense set of quantiles or using small dataset~\cite{kneib2013beyond}. Various methods have been proposed in the literature to overcome this problem~\cite{dette2008non,schnabel2013simultaneous} but they always require additional efforts and are not always applicable~\cite{kneib2013beyond}. Furthermore, one can use a collection of point forecasts for a specific quantity or event as an ensemble model for probabilistic forecasting. In this setup, we need some form of statistical post-processing~\cite{gneiting2014probabilistic}. State-of-the-art techniques for statistical post-processing include the non-homogeneous regression (NR) or ensemble model output statistics (EMOS) technique proposed by Gneiting et al.~\cite{gneiting2005weather} and the ensemble Bayesian model averaging (BMA) approach developed by Raftery et al.~\cite{raftery2005using}. For an in-depth review of probabilistic forecasting, please refer to~\cite{gneiting2014probabilistic}.

Besides these methods, researchers employ Bayesian probability theory to provide approaches for probabilistic forecasting. Bayesian probability theory offers mathematically grounded tools to reason about model uncertainty, but these usually come with a prohibitive computational cost~\cite{gal2016dropout}. This computation complexity stems from marginalising, a computationally intensive integration required by Bayesian models which makes the computation for complex models impossible. To solve this problem, many approximate integration algorithms have been developed, including Markov chain Monte Carlo (MCMC) methods, variational approximations, expectation propagation, and sequential Monte Carlo~\cite{neal1993probabilistic,jordan1999introduction,doucet2001introduction,minka2001expectation}. However, this method still suffers from a prohibitive computational cost, rapid growth in the number of parameters and time-intensive convergence~\cite{gal2016dropout}. Furthermore, the success of Bayesian model heavily relies on selecting a prior distribution. Selecting a suitable prior distribution is a delicate task which requires insight into the data. Recently, Gal et al.~\cite{gal2016dropout} use dropout~\cite{srivastava2014dropout} layers for probabilistic machine learning. While dropout is used in many models in deep learning as a way to avoid over-fitting, Gal has shown in his paper that a neural network with arbitrary depth and non-linearities, with dropout applied before every weight layer, is mathematically equivalent to an approximation to the probabilistic deep Gaussian process~\cite{damianou2013deep}. For an in-depth review of Bayesian probabilistic machine learning, please refer to~\cite{ghahramani2015probabilistic,gal2016uncertainty}.

\subsection{Generative Adversarial Networks}
Generative Adversarial Network (GAN)~\cite{goodfellow2014generative} is a new type of neural networks which enables us to learn an unknown probability distribution from samples of the distribution. GAN can learn the probability distribution of a given dataset and generate synthetic data which follows the same distribution. As a result, they are capable of synthesizing artificial data which looks realistic. While GANs were originally proposed to solve the problem of data scarcity, its promising results have drawn a lot of attention in the research community and many interesting derivations, extensions, and applications have been proposed for GANs~\cite{DBLP:journals/corr/RadfordMC15,arjovsky2017wasserstein,DBLP:journals/corr/DonahueKD16,DBLP:journals/corr/ChenDHSSA16,mirza2014conditional}. Unfortunately, despite their remarkable performance, evaluating and comparing GANs is notoriously hard. Thus, the application of GANs is limited to the domains where the results are intuitively assessable like image generation~\cite{DBLP:journals/corr/RadfordMC15}, music generation~\cite{mogren2016c}, voice generation~\cite{gao2018voice}, and text generation~\cite{yu2017seqgan}.\bigskip

Diverse approaches have been proposed for probabilistic forecasting. However, each of these methods has limitations which prevent them from becoming canonical approach in the industry. Hence, mean regression methods are widely employed by industry section with their critical shortcomings. As mentioned before, GANs are a powerful method for learning probability distributions. In this paper, we exploit the potentials of GANs to learn full probability distribution of future values without the restrictions of the aforementioned methods. We introduced ForGAN, a conditional GAN~\cite{mirza2014conditional} for probabilistic forecasting. The main contributions of this paper are as follows:
\begin{itemize}
	\item We propose ForGAN, a novel approach to employ a conditional GAN for forecasting future value. Our method can learn the full conditional probability distribution of future values even in complex situations without facing conventional problems of probabilistic forecasting methods such as quantile crossing or dependency on the chosen prior distribution.
	\item We conduct various experiments to investigate the predictive capabilities of our method and compare it with the state-of-the-art methods as well as a conventional regression neural network model with a similar structure to ForGAN. Our method outperforms its counterparts on various metrics.
	\item We introduce a new dataset for later reference and comparison.
\end{itemize}

\section{Related work}
\label{sec:related_work}

Lately, GANs have been applied to various problems in the sequential data domain and achieved remarkable results. In this section, we give a brief overview of studies related to our work.

Most research regarding applying GAN on sequential data is concerned with discrete problems, e.g. text generation task. Since the discrete space of words cannot be differentiated in mathematics, modifying a GAN to work with discrete data is a challenging task. Many papers have been published to address this problem and they have reported remarkable results~\cite{press2017language,li2017adversarial,yu2017seqgan,zhang2016generating}. However, we are interested in the (quasi) continues regime. Therefore, these techniques are not directly applicable here.

In the continuous regime, we find GANs being utilized to generate auditory data. C-RNN-GAN~\cite{mogren2016c} works on music waveforms as continuous sequential data to generate polyphonic music. This GAN uses  Bidirectional LSTM in the structure of the generator and discriminator.  Moreover, there are many other studies on auditory data which work on audio spectrograms and consider them as 2D images. For instance, Donahue et al.~\cite{donahue2018exploring} as well as Michelsanti, Tan et al.~\cite{michelsanti2017conditional} employ GAN on audio spectrograms for speech enhancement. Fan et al.~\cite{fan2018svsgan} propose a GAN for separating the singing voice from background music. Donahue et al.~\cite{DBLP:journals/corr/abs-1802-04208} propose a GAN for synthesizing raw-waveform audio and Gao et al.~\cite{gao2018voice} employ GAN for synthesizing of impersonated voices. However, contrary to our work these studies are first not concerned with forecasting and second, the results are intuitive. The latter point is important as analogous to the image domain, music can be judged by listening to it. 

Since there is no consensus on a process for evaluating GANs,  application of GANs beyond text and auditory data is a very challenging task. We found a few attempts on the application of GANs beyond these data types. Hyland and Esteban~\cite{esteban2017real} propose RGAN and RCGAN to produce realistic real-valued multi-dimensional medical time series. Both of these GANs employ LSTM in their generator and discriminator while RCGAN uses Conditional GAN instead of Vanilla GAN to incorporate a condition in the process of data generation. 
They also describe novel evaluation methods for GANs, where they generate a synthetic labeled training dataset and train a model using this set. Then, they test this model using real data. They repeat the same process using a real train set and synthetic labeled test set. GAN-AD~\cite{li2018anomaly} is proposed to model time-series for anomaly detection in Cyber-Physical Sytems (CPSs). This GAN uses LSTM in both generator and discriminator, too. Zhang et al.~\cite{zhanggenerative} propose a conditional GAN for generating synthetic time-series in smart-grids. Unlike previous work, this GAN employs CNN to construct generator and discriminator.

To the best of our knowledge, this is for the first time that a GAN is employed for the forecasting task. Our work is analogous to RCGAN~\cite{esteban2017real}, however, we pursue a different goal. As a result, we need to take a different approach to train and evaluate the performance of ForGAN.

\section{Methodology}
\label{sec:method}
\subsection{Generative Adversarial Network (GAN)}
Generative Adversarial Networks (GANs)~\cite{goodfellow2014generative} are a class of algorithms for modeling a probability distribution given a set of samples from the data probability distribution $\rho_\text{data}$. A GAN consists of two neural networks namely \textbf{generator} $G$ and \textbf{discriminator} $D$. These components are trained simultaneously in an adversarial process. First, a noise vector $z$ is sampled from a known probability distribution $\rho_\text{noise}(z)$ (normally a Gaussian distribution). $G$ takes the noise vector $z$ as an input and trains to generate a sample whose distribution follows $\rho_\text{data}$. On the other hand, $D$ is optimized to distinguish between generated data and real data. In other words, $D$ and $G$ play the following two-player minimax game with value function $V (G, D)$:
\begin{equation*}
\begin{aligned}
\min_{G} \max_{D} V(D,G) = \ &\mathbb{E}_{x\sim\rho_\text{data}(x)}[\text{log}\, D(x)] + \\ &\mathbb{E}_{z\sim\rho_\text{noise}(z)}[\text{log}\,(1-D(G(z)))]\,.
\label{eq:gan}
\end{aligned}
\end{equation*}
While training the GAN, generator $G$ learns to transform a known probability distribution $\rho_{z}$ to the generators distribution $\rho_{G}$ which resembles $\rho_\text{data}$.

\subsection{Conditional GAN (CGAN)}
Conditional GAN (CGAN)~\cite{mirza2014conditional} is an extension of GAN which enables us to condition the model on some extra information $y$. This could be any kind of auxiliary information, such as class labels or data from other modalities. We can perform the conditioning by feeding $y$ into both the discriminator and generator as additional input layer. The new value function $V(G,D)$ for this setting is:

\begin{equation*}
\begin{aligned}
\min_{G} \max_{D} V(D,G) = \ &\mathbb{E}_{x\sim\rho_\text{data}(x)}[\text{log}\, D(x|y)] +\\ &\mathbb{E}_{z\sim\rho_{z}(z)}[\text{log}\,(1-D(G(z|y)))]\,.
\label{eq:cgan}
\end{aligned}
\end{equation*}

\subsection{Probabilistic forecasting with CGAN}

\begin{figure*}
	\centering
	\includegraphics[width=0.94\textwidth]{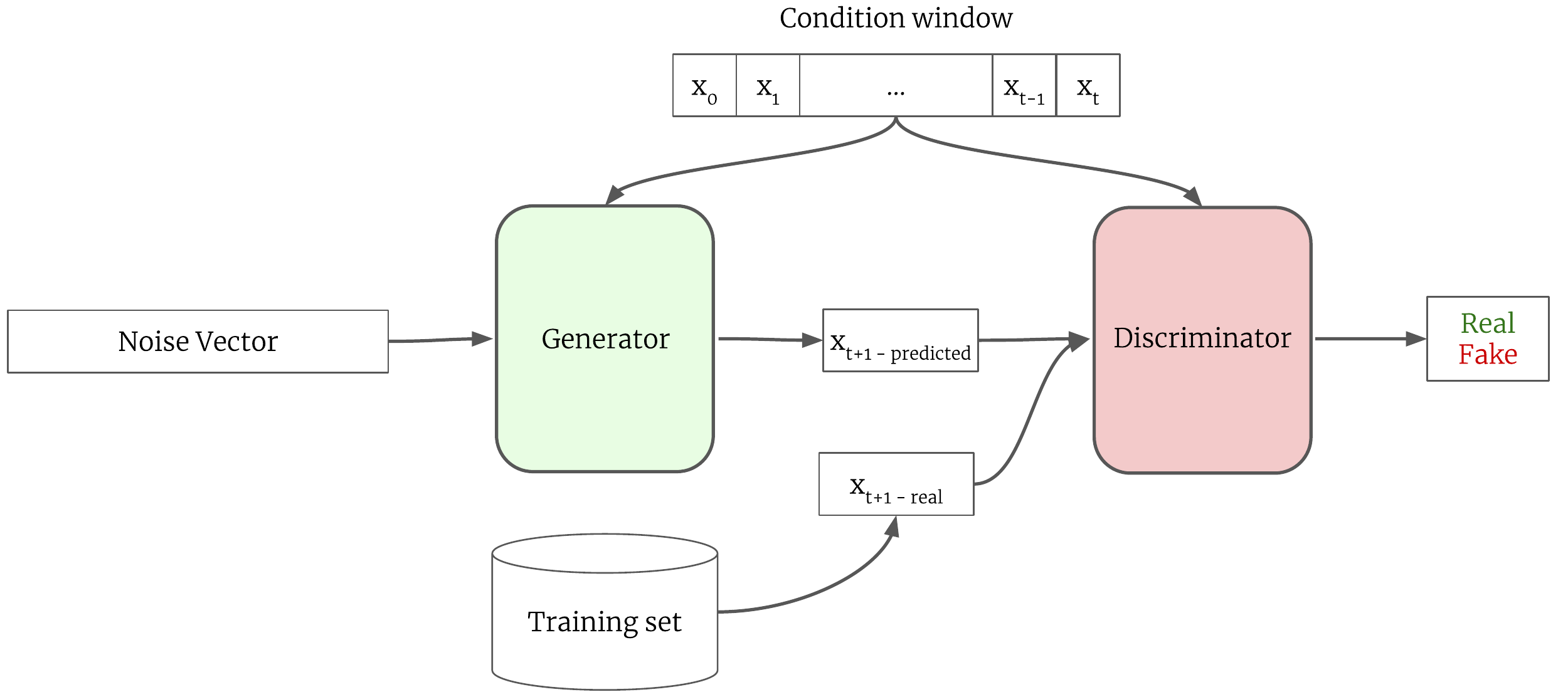}
	\caption{Overview of proposed ForGAN architecture. The condition $c$ is handed to generator $G$ and discriminator $D$.  }
	\label{fig:gan}
\end{figure*}

In this paper we aim to model the probability distribution of one step ahead value $x_{t+1}$ given the historical data $c=\{x_0,.., x_t\}$, i.e. $\rho(x_{t+1}|c)$. We employ CGAN to model $\rho(x_{t+1}|c)$. Figure~\ref{fig:gan} presents an overview of ForGAN. The historical data is provided to generator and discriminator as condition. The generator takes the noise vector which is sampled from a Gaussian distribution with mean $0$ and standard deviation $1$ and forecasts $x_{t+1}$ with regard to the condition window $c$. The discriminator takes the $x_{t+1}$ and inspects whether it is a valid value to follow $c$ or not. Hence, the ForGAN value function is:

\begin{equation*}
\begin{aligned}
\min_{G} \max_{D} V(D,G) = \ &\mathbb{E}_{x_{t+1}\sim\rho_\text{data}(x_{t+1})}[\text{log}\, D(x_{t+1}|c)] +\\ &\mathbb{E}_{z\sim\rho_{z}(z)}[\text{log}\,(1-D(G(z|c)))]
\label{eq:lose_forgan}
\end{aligned}
\end{equation*}

By training this model, the optimal generator models the full probability distribution of $x_{t+1}$ for a given condition window. With having the full probability distribution in hand, we can extract information regarding any possible outcome and the probability of their occurrence by sampling.

\subsection{Architecture}

\begin{figure*}[t]
	\centering
	\subfloat[\label{fig:gen}]{%
		\includegraphics[width=0.7\linewidth]{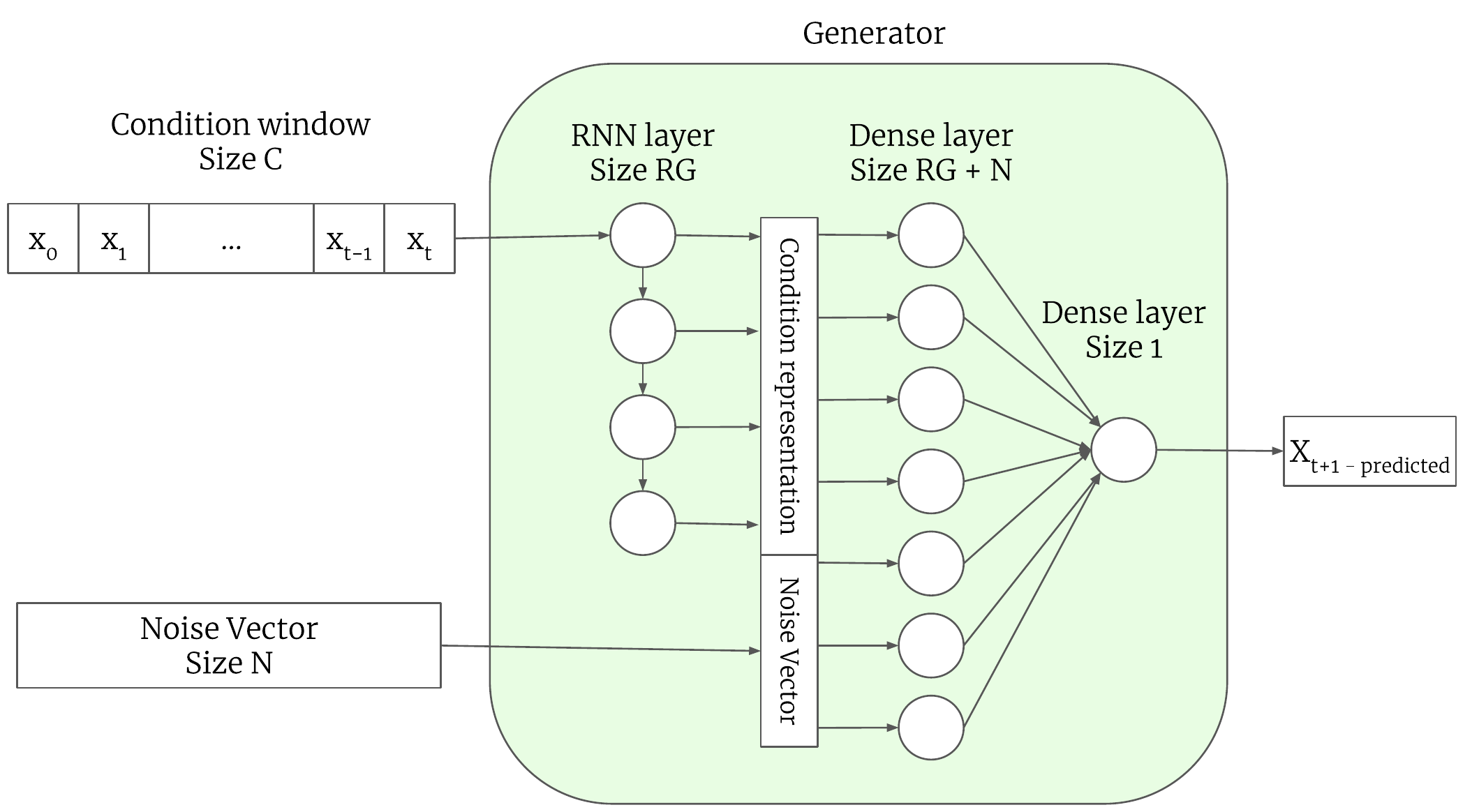}}
	\\
	\subfloat[\label{fig:dis}]{%
		\includegraphics[width=0.7\linewidth]{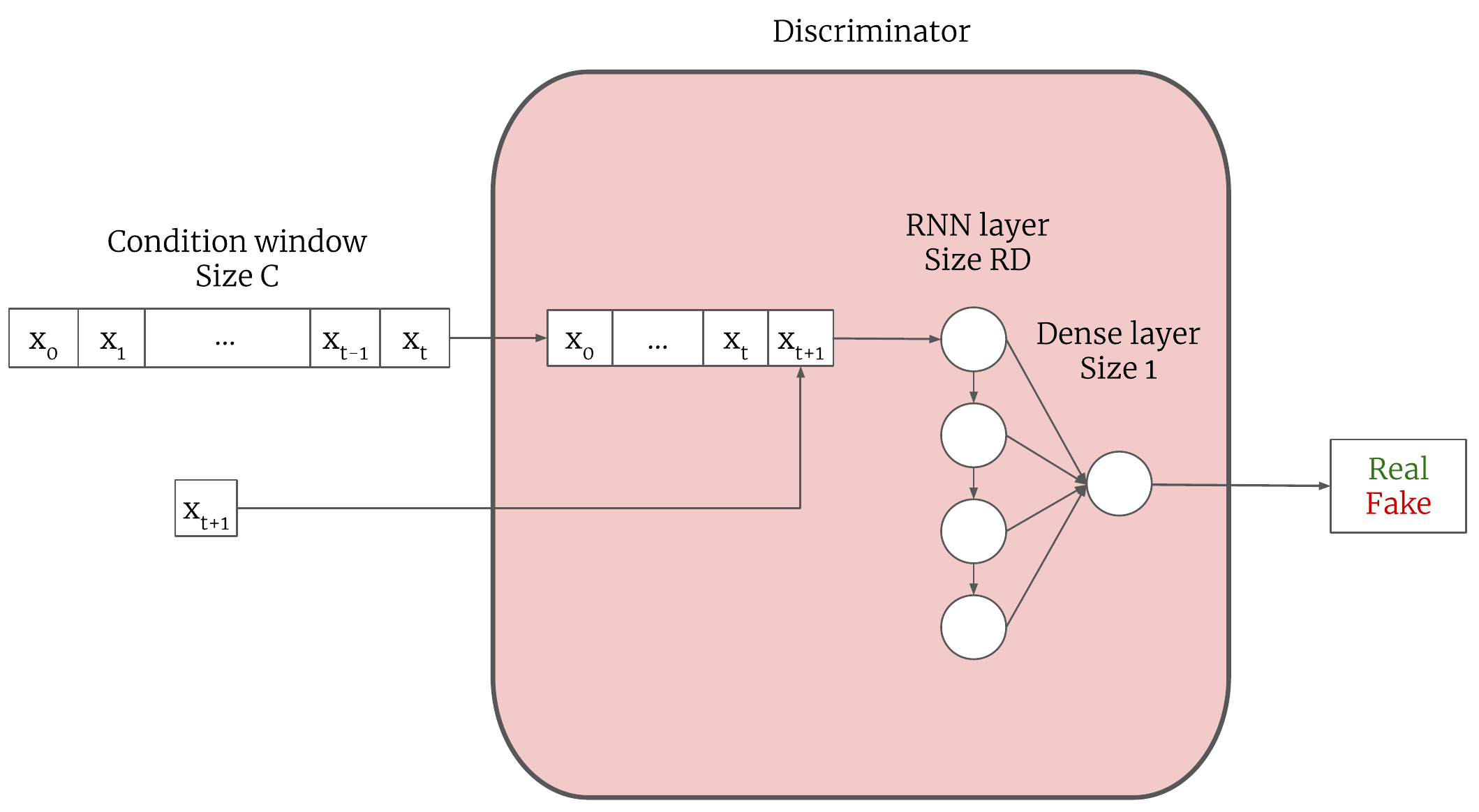}}
	\caption{(a): The architecture of the generator in detail. The generator takes noise vector and a time windows and forecasts the value of next step ($x_{t+1}$) . (b): The architecture of the discriminator in detail. The discriminator receives $x_{t+1}$ and time window and determines if $x_{t+1}$ is valid.}
	\label{fig:gen_dis} 
\end{figure*}

We use one of the members of RNN family as the main component of both generator and discriminator. We select between LSTM or GRU using the procedure described in section~\ref{sec:hp_tune}. The generator (Fig.~\ref{fig:gen_dis}\subref{fig:gen}) takes the condition window and passes the condition through an RNN layer to construct its representation. Then, it concatenates the condition representation with the noise vector and passes them through two dense layers which result in the predicted $x_{t+1}$ value. The discriminator (Fig.~\ref{fig:gen_dis}\subref{fig:dis}) takes $x_{t+1}$ either from the generator or the dataset alongside the corresponding condition  window and concatenates $x_{t+1}$ at the end of the condition window to obtain $\{x_0,.., x_{t+1}\}$. The rest of the network tries to check the validity of this time window. For this purpose, it passes the obtained time window through an LSRM/GRU layer followed by a dense layer to acquire a single value which specifies the validity of the  aforementioned time window.

\subsection{G-regression model}

Normally, forecasting models are trained by optimizing a point-wise error metric as loss function however we employ adversarial training to train neural network for forecasting. To study the effectiveness of adversarial training in comparison to conventional training, we construct the G-regression model, a model with identical structure to generator $G$.To follow the conventional way of training neural networks for forecasting, we train this model by optimizing RMSE as the loss function and compare its results with ForGAN.

\section{Experiments}
\label{sec:experiments}
To investigate the performance of ForGAN, we test our method with three experiments and later on, if applicable, compare results with the state-of-the-art methods. Let us first introduce the used datasets. Next, we elaborate on common evaluation methods for forecasting task and how we assess ForGAN to produce correct and meaningful results. We then demonstrate how we chose the particular hyperparameters for each dataset. Last but not least we elaborate on the setup of our experiments.

\subsection{Datasets}
\subsubsection{Lorenz Dataset}

\begin{figure*}[t]
	\centering
	\subfloat[\label{fig:lorenz_1}]{%
		\includegraphics[width=0.45\linewidth]{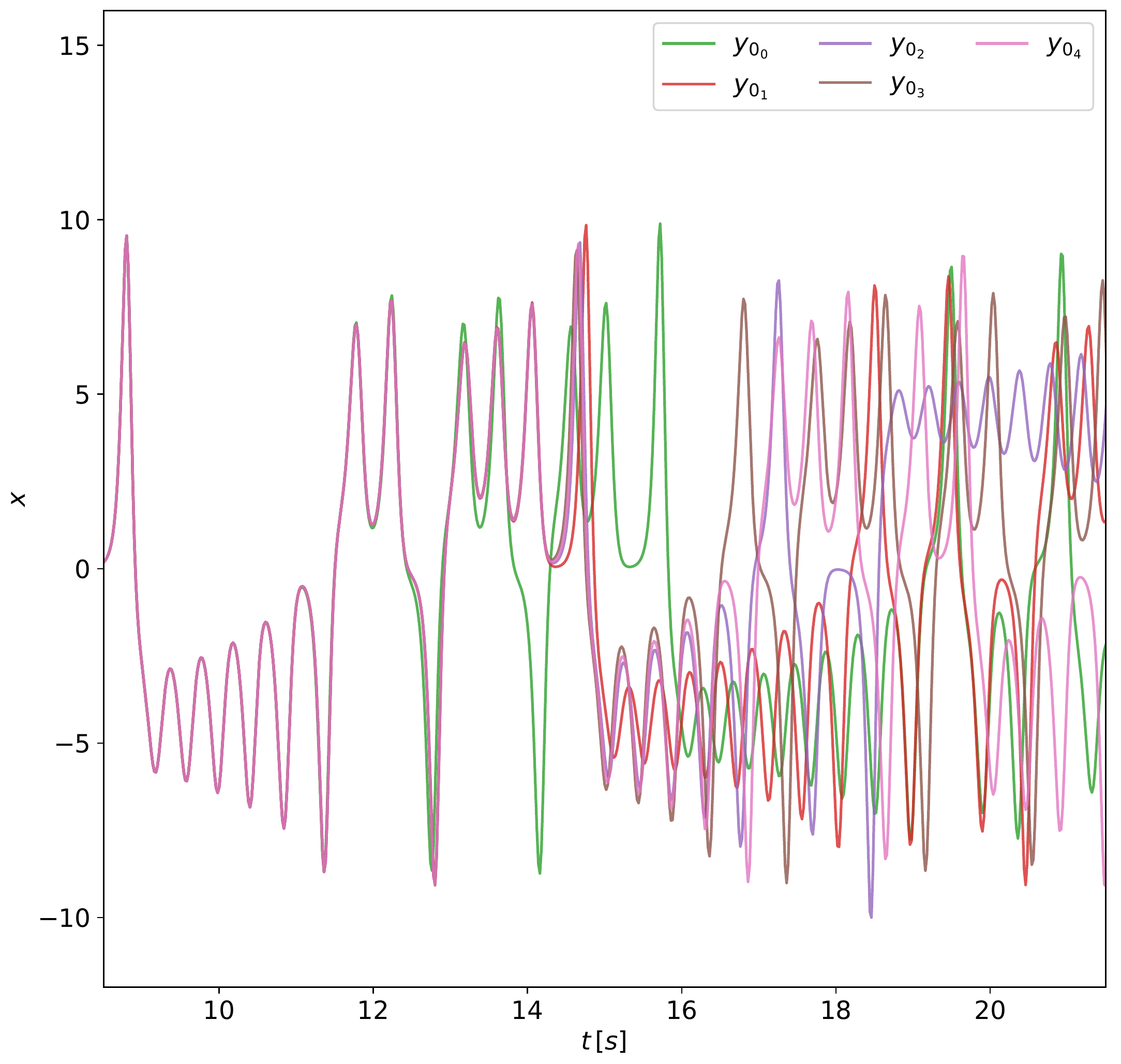}}
	\hfill
	\subfloat[\label{fig:lorenz_2}]{%
		\includegraphics[width=0.45\linewidth]{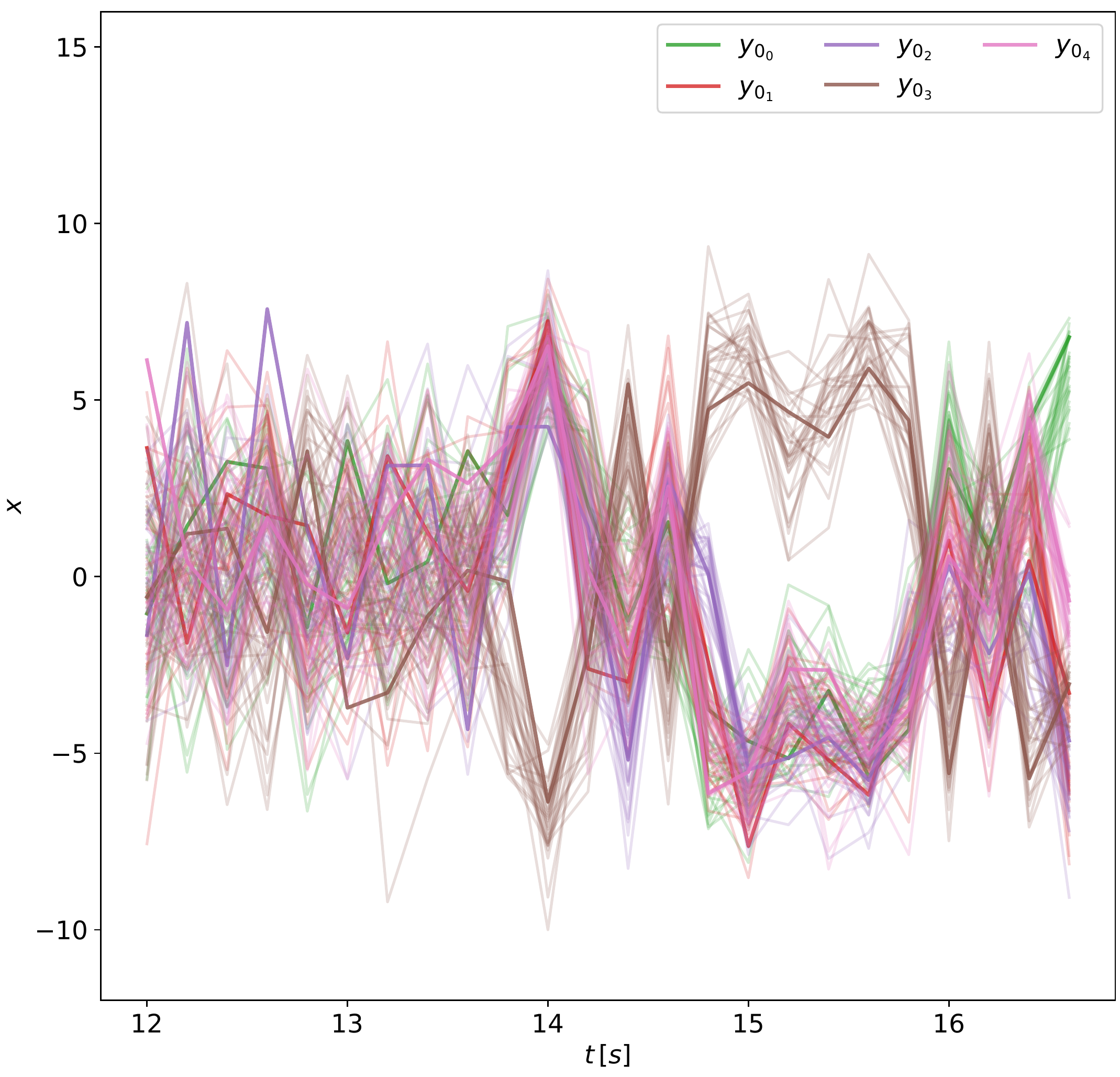}}
	\\
	\subfloat[\label{fig:lorenz_3}]{%
	\includegraphics[width=0.45\linewidth]{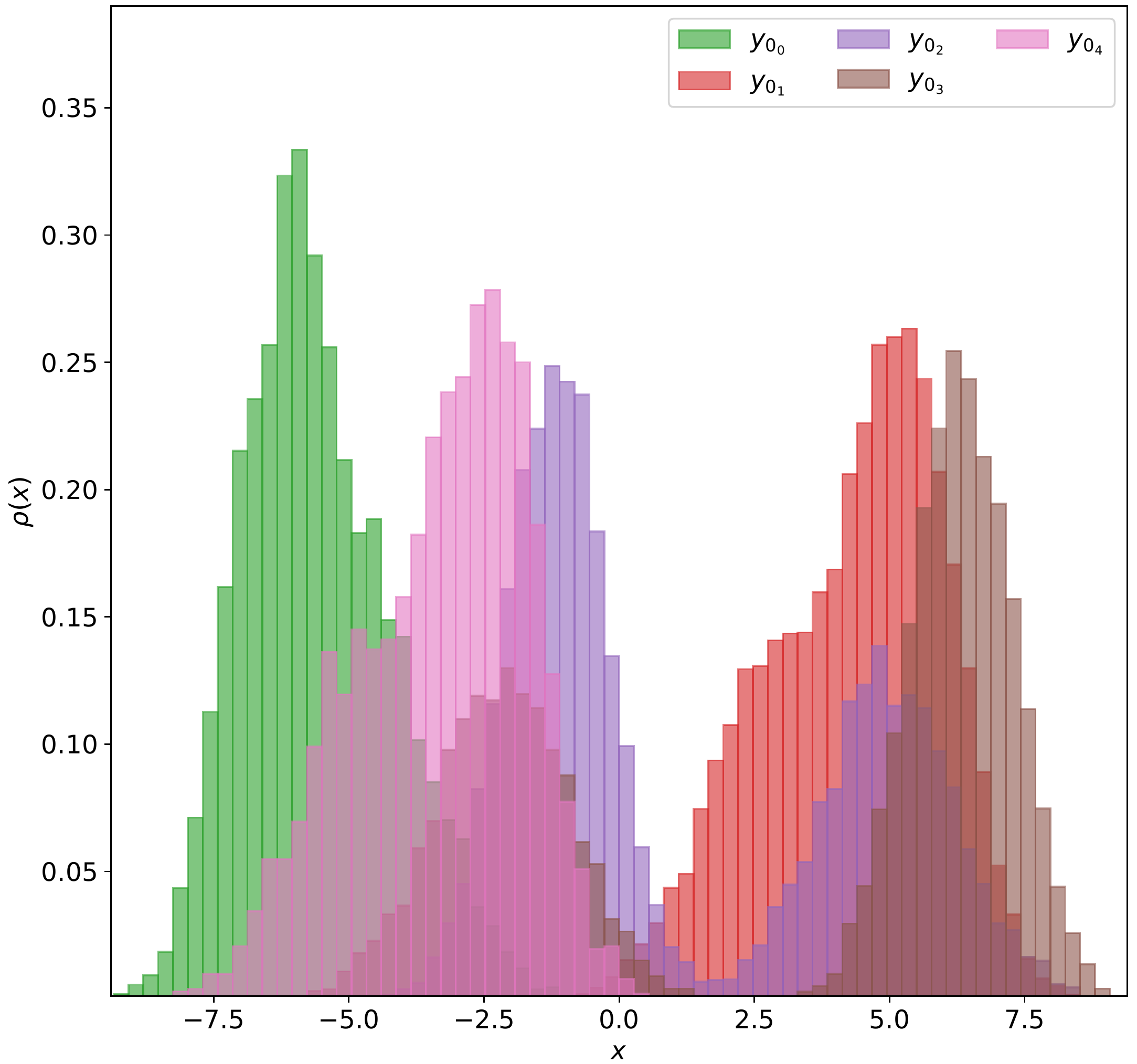}}
	\hfill
	\subfloat[\label{fig:lorenz_4}]{%
		\includegraphics[width=0.45\linewidth]{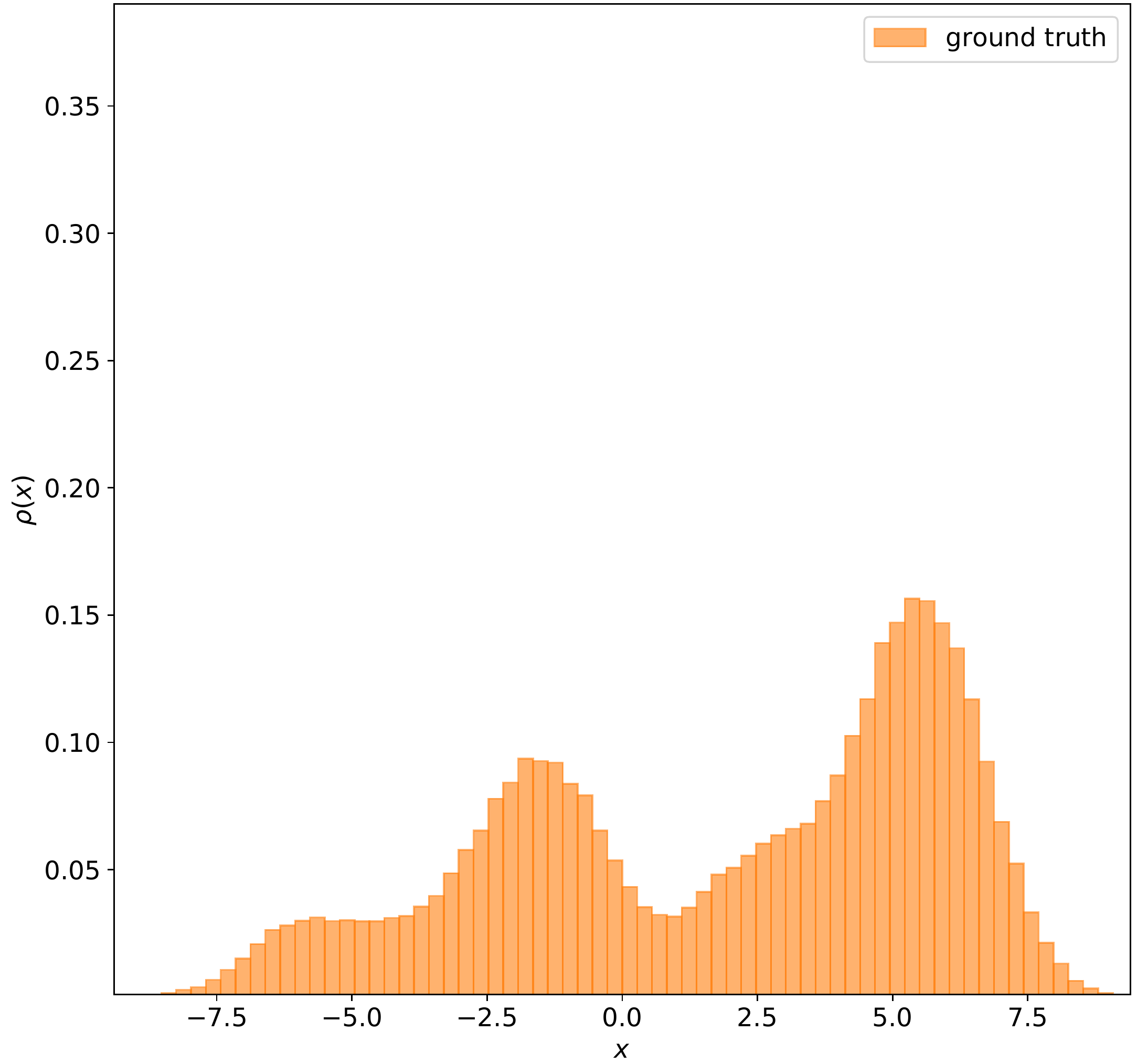}}
	\caption{(a): Solution to the Lorenz system for different initial values $y_0$. (b): The bifurcation region after the data augmentation steps described in the text. (c): Possible values $x_{t+1}$ distinguished by initial value $y_0$. (d): Full probability distribution of $x_{t+1}$.}
	\label{fig:lorenz} 
\end{figure*}

\begin{table}
	\centering
	\caption{Initial values $y_0$ and relative composition of our Lorenz dataset.}
	\renewcommand{\arraystretch}{1.5}
	\begin{tabular}{clc}
		\toprule
		\textbf{Index} & $y_0$ & \textbf{Relative Occurrence} \\
		\midrule
		0 & 1.0001         & 5.5 \% \\
		1 & 1.000001       & 22 \%  \\
		2 & 1.00000001     & 42 \%  \\
		3 & 1.0000000001   & 24 \%  \\
		4 & 1.000000000001 & 6.5 \% \\
		\bottomrule
	\end{tabular}
	\label{tab:composition}
\end{table}
In the first experiment, we create a complex dataset to inspect the probabilistic forecasting capability of our method.  We form a dataset which contains multiple time window clusters. Each cluster consists of similar and complex time windows generated using the Lorenz system. The Lorenz equations describe the atmospheric convection $x$, the horizontal temperature variation $y$, and the vertical temperature $z$ as a function of time $t$. Using a dot for temporal derivative the system of coupled differential equations is given by
\begin{align}
\dot{x} &= \sigma (y-x)\,, \notag\\
\dot{y} &= x ( \rho -z)\,, \notag\\
\dot{z} &= y\,x - \beta\,z\,, 
\label{eq:lorenz}
\end{align}
where $\sigma$ is proportional to the Prandtl number~\cite{white2006viscous}, $\rho$ is proportional to the Rayleigh number~\cite{chandrasekhar2013hydrodynamic} and $\beta$ is connected to physical dimensions of the atmospheric layer of interest~\cite{sparrow2012lorenz}. One of the most interesting features of the Lorenz equations is the emergence of chaotic behavior~\cite{kellert1993wake,sparrow2012lorenz} for certain choices of the parameters $\sigma$, $\rho$, and $\beta$. In the following we fix $\sigma=16$, $\rho = 45.92$, and $\beta = 4$. Furthermore, we fix the initial conditions $x_0=1$ and $z_0=1$. To construct the dataset, first, we select five $y_0$ to serve as the seeds for our clusters and specify the relative occurrence of clusters as presented in Tab.~\ref{tab:composition}. Then we generate 100000 data samples with the length of 26 seconds and the resolution of 0.02s using these $y_0$. The result is presented in Fig.~\ref{fig:lorenz}\subref{fig:lorenz_1}. We add a Gaussian noise with mean $0$ and standard deviation of $7.2$ to create unique time windows while preserving similarity inside each cluster. From Fig.~\ref{fig:lorenz}\subref{fig:lorenz_1}, we locate the bifurcation region and select the region between 12 and 17 seconds as the condition time window for training. The dataset of condition time windows is plotted in Fig.~\ref{fig:lorenz}\subref{fig:lorenz_2}. Finally, for the target values $x_{t+1}$, we sample randomly from $t\in(20, 22,25)$ which forms the probability distributions as they are presented in Fig.~\ref{fig:lorenz}\subref{fig:lorenz_3}. Fig.~\ref{fig:lorenz}\subref{fig:lorenz_4} presents the full probability distribution of the $x_{t+1}$ for the entire dataset.

\subsubsection{Mackey-Glass Dataset}

The time delay differential equation suggested by Mackey and Glass~\cite{mackey1977oscillation} has been used widely as a standard benchmark model to generate chaotic time-series for the forecasting task.
\begin{equation}
\label{eq:mg}
\dot{x} = \frac{a\,x(t - \tau)}{(1 + 10\cdot(t - \tau)) - b\,x(t)} \,.
\end{equation}
To make our result comparable with state-of-the-art~\cite{mendez2017competitive}, we set $a = 0.1$, $b = 0.2$ and $\tau = 17$. We generate a dataset with length 20000 using Eq.(~\ref{eq:mg}) for our second experiment. 

\subsubsection{Internet Traffic Dataset}
For our last experiment, we apply our method to a real-world problem, forecasting internet traffic. We use a dataset which belongs to a private ISP with centers in eleven European cities (which is commonly known as A5M)~\cite{cortez2012multi}. It contains data corresponding to a transatlantic link and was collected in 2005 from 06:57 on 7th of June to 11:17 on 29th of July.

\subsection{Evaluation Metrics}

Commonly, in forecasting tasks, point-wise error metrics are used. To be able to compare to the state-of-the-art we report RMSE, MAE and MAPE which are related to each other by
\begin{align}
\text{RMSE}  &= \sqrt{\frac{1}{N} \sum\limits_i (x_i - \hat{x}_i)^2}\,,
\label{eq:rmse} \\
\text{MAE}   &= \frac{1}{N} \sum\limits_i \left| x_i - \hat{x}_i \right|\,,
\label{eq:mae} \\
\text{MAPE}  &= \frac{1}{N} \sum\limits_i \left| 10^2 \times \frac{ x_i - \hat{x}_i }{x_i} \right|\,.
\label{eq:mape}
\end{align}
Here $N$ is the number of data samples $x_i$, and $\hat{x}_i$ are the actual predictions. However, 
point-wise error metrics are not suitable for assessing distributions similarities. Since ForGAN models the full probability distribution of $x_{t+1}$, we are interested in measuring how accurate we managed to reproduce the data distribution. Therefore, we select the Kullback-Leibler divergence (KLD)~\cite{kullback1951} to report the performance of our method. KLD measures the divergence between two probability distributions $P$ and $Q$. Since we have finite data samples and ForGAN by nature samples, we select the discrete version of KLD which is defined as:
\begin{align}
\label{eq:kl}
\text{KL} \left( P|Q \right) &= \sum\limits_i P_i\, \text{log}\, \frac{P_i}{Q_i}\,.
\end{align}
Note, $P$ denotes data distribution and $Q$ indicates prediction probability distribution. Hence, due to the appearance of $Q$ in the denominator, if predictions distribution does not cover data distribution correctly KLD is not defined. To determine the optimal number of bins for the histogram of distribution, we follow the method suggested in~\cite{knuth2006optimal} which aims for the optimum between shape information and noise. 
To evaluate our method and compare our results to the state-of-the-art in one step ahead forecasting, we train ForGAN alongside G-regression and report RMSE, MAE, MAPE, and (if possible) KLD. To compute KLD for ForGAN, we sample $100$ forecast of $x_{t+1}$ for any condition in the test set. Then, we form ForGAN's prediction probability distribution for the entire test set and calculate KLD between this distribution and test set data distribution. Therefore, the G-regression model does not output probability distribution. Thus, we use the histogram of G-regression predictions to calculate KLD. To calculate point-wise error metrics for ForGAN, we run it $100$ times over the test set and report the mean and the standard deviation of these metrics as the result. With this information, we present a complete and clear image of ForGAN performance. The KLD value shows how accurate our method has learned the data distribution and the point-wise error metrics specifies how well it has considered the condition to forecast $x_{t+1}$. Furthermore, we have the possibility to compare ForGAN with other methods based on various criteria.
\subsection{Hyperparameter Tuning}
\label{sec:hp_tune}

The ForGAN structure has a set of hyperparameters which we tune for each experiment separately. Tab.~\ref{tab:hyper} provides the list of hyperparameters alongside the allowed values. During training, at each epoch, we train the discriminator several times but our generator is only trained once per epoch. The number of training iterations for discriminator in one epoch is one of the hyperparameters  ($\text{D}_\text{Iter}$ in Tab.~\ref{tab:hyper}).
\begin{table*}
	\centering
	\caption{The List of ForGAN hyperparameters alongside the range of allowed values.}
	\begin{tabular}{rcc}
		\toprule
		\textbf{Hyperparameters}                       & \textbf{Abbreviation}      &   \textbf{Values}\\
		\midrule
		The type of cells                          & T                       & GRU, LSTM\\
		The number of cells in generator           & RG                       & 1, 2, 4, 8, 16, 32, 64, 128, 256\\
		The number of cells in discriminator       & RD                      & 1, 2, 4, 8, 16, 32, 64, 128, 256\\
		The size of noise vector                       & N                      & 1, 2, 4, 8, 16, 32\\
		The size of look-back window ( Condition )     & C                     & 1, 2, 4, 8, 16, 32, 64, 128, 256\\
		Number of training iteration for discriminator & $\text{D}_\text{Iter}$ & 1, 2, 3, 4, 5, 6, 7\\
		\bottomrule
	\end{tabular}
	\label{tab:hyper}
\end{table*}
To tune these hyperparameters, we use a genetic algorithm. Genetic algorithms~\cite{holland1992adaptation} are a class of methods for optimization task which are inspired by the evolution in nature. These methods provide an alternative to traditional optimization techniques by using directed random searches to locate optimal solutions in complex landscapes~\cite{srinivas1994genetic}. The hyperparameters are encoded in a vector which is called a gene. The algorithm starts with a set of randomly initialized genes to form a gene pool and tries to find the most optimized gene through iterative progress. In each iteration, the genes in the gene pool are evaluated using a fitness function and those with low scores are eliminated. Then, the remaining genes are used to create offsprings. After multiple iterations, the algorithm converges to a gene with the most optimized combination of values. For further detailed information on genetic algorithms, we refer to this comprehensive survey~\cite{srinivas1994genetic}.

Our genetic algorithm has a gene pool of size 8 and we run it for 8 iterations. At each iteration, we use 4 of the genes with the best scores to create offsprings. 4 new genes are created using crossover while 4 other genes are created using mutation. Using the values of a gene, we construct a ForGAN and train it on a train set while we monitor KLD on a validation set.

\subsection{Setup}

For each experiment, the dataset is divided into three subsets. 50\% of the dataset is used as the train set, 10\% as the validation set and 40\% as the test set. We code the ForGAN using TensorFlow~\cite{tensorflow2015-whitepaper} and run it on a DGX-1 machine.


\section{Results and Discussion}
\label{sec:results}

In Tab.~\ref{tab:hyper_xp} we present the set of optimal hyperparameters which are found by the genetic algorithm for each experiment. The numerical results achieved by ForGAN are summarized in Tab.~\ref{tab:all_res} alongside the state-of-the-art results on the Mackey-Glass dataset~\cite{mendez2017competitive} and Internet traffic dataset (A5M)~\cite{cortez2012multi}. Furthermore, we report the results obtained from G-regression model.
\begin{table*}
	\centering
	\caption{The optimal hyperparameters used to construct ForGAN for different experiments.}
	\begin{tabular}{c|cccc}
		\toprule
		\multicolumn{2}{c}{ } & \textbf{Lorenz} & \textbf{Mackey-Glass} & \textbf{Internet Traffic Data} \\
		\multicolumn{2}{c}{ } & \textbf{Experiment} & \textbf{Experiment} & \textbf{Experiment} \\
		\midrule
		& T  & GRU & LSTM & GRU \\
		&RG & 8 & 64 & 8 \\
		Hyperparameters  &RD & 64 & 256 & 128 \\
		& N & 32 & 4 & 16 \\
		& C & 24 & 32 & 32 \\
		& $\text{D}_\text{Iter}$ & 2 & 6 & 3 \\
		\bottomrule
	\end{tabular}

	\label{tab:hyper_xp}
\end{table*}
\begin{table*}
	\caption{The results achieved by ForGAN alongside the results from G-regression model and state-of-the-art on Mackey-Glass dataset~\cite{mendez2017competitive} and Internet traffic dataset~\cite{cortez2012multi}. The numbers in the parenthesis indicate the one standard deviation of results.}
	\label{tab:all_res}
	\small 
	\centering
	\begin{tabular}{c|ccllcl}
		\toprule
		\multicolumn{2}{c}{ }& \textbf{state-of-the-art} &\textbf{G-regression} &\multicolumn{3}{c}{ForGAN [\textbf{Our Method}]}\\
		\midrule
		&RMSE	& -    &$\mathbf{2.91}$	&$4.06\,(0.01)$	 \\
		Lorenz	&MAE    & -    &$\mathbf{2.39}$	        &$2.94\,(0.01)$	 \\
		dataset &MAPE	& -    &$\mathbf{2.25\,\%}$	    &$3.35\,(0.24)\, \%$ \\
		&KLD    & -    &Nan 	        &$\mathbf{1.67\times10^{-2}}$	     \\
		
		\multicolumn{2}{c}{ }\\
		
		&RMSE    & $4.38\times10^{-4}$    &$5.63\times10^{-4}$        &$\mathbf{3.82\,(0.02)\times10^{-4}}$    \\
		Mackey-Glass &MAE     & -                      &$4.92\times10^{-4}$        &$\mathbf{2.93\,(0.01)\times10^{-4}}$    \\
		dataset      &MAPE    & -                      &$6.29\times10^{-2}\,\%$    &$\mathbf{3.46\,(0.02)\times10^{-2}\,\%}$ \\
		&KLD     & -                      &$8.00\times10^{-3}$        &$\mathbf{3.18\times10^{-3}}$    \\
		
		\multicolumn{2}{c}{ }\\
		
		&RMSE    & -          &$\mathbf{1.27\times10^{8}}$  &$1.31\,(0.00)\times10^{8}$  \\
		Internet traffic    &MAE     & -          &$\mathbf{9.01\times10^{7}}$  &$9.29\,(0.03)\times10^{7}$  \\
		dataset (A5M)       &MAPE    &$2.91\,\%$  &$\mathbf{2.85\,\%}$ &$2.94\,(0.01)\,\%$  \\
		&KLD     & -          &$5.31\times10^{-11}$               &$\mathbf{2.84\times10^{-11}}$ \\
		
		\bottomrule
	\end{tabular}
\end{table*}
\medskip

\begin{figure*}
	\centering
	\includegraphics[width=\textwidth]{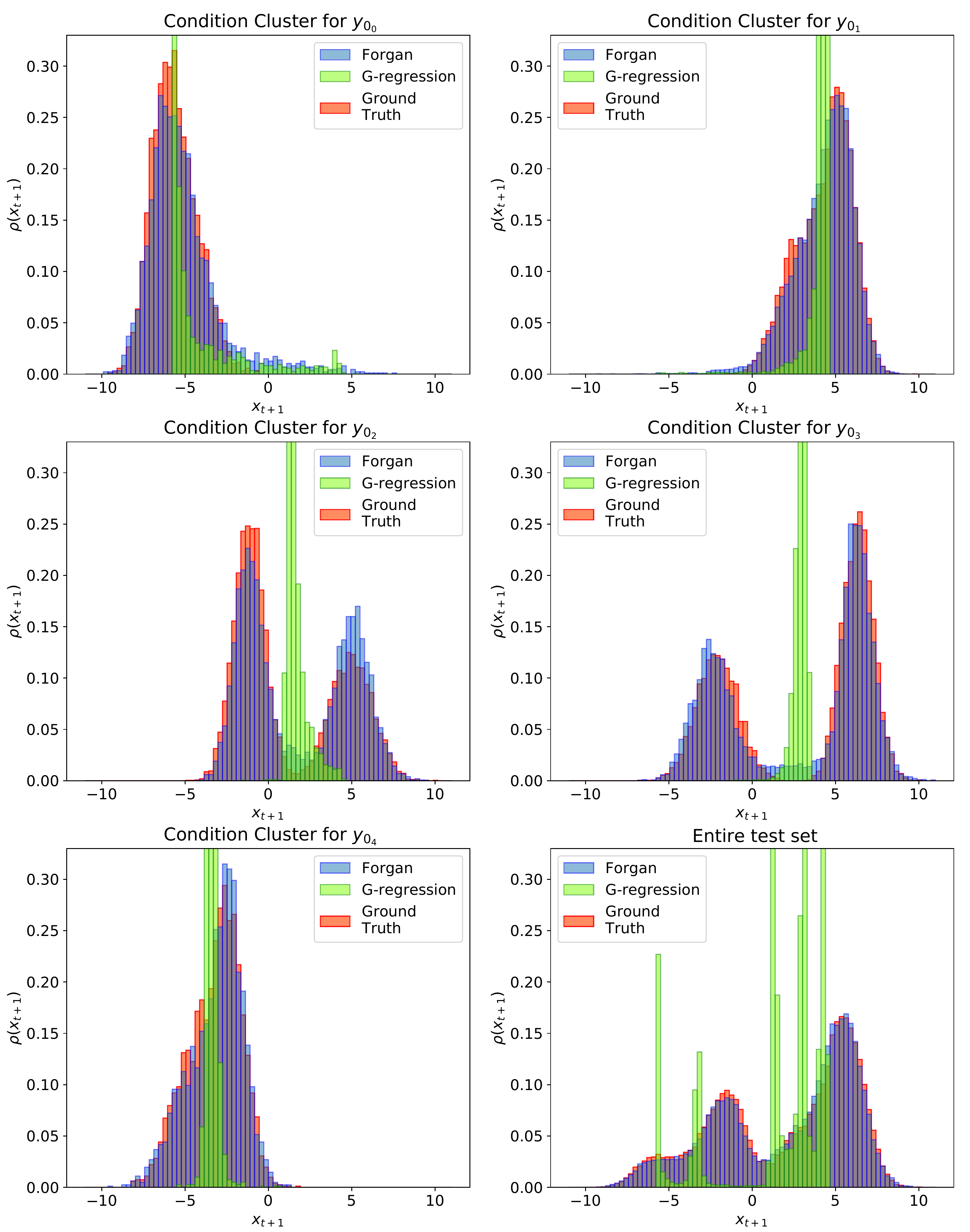}
	\caption{The prediction of $x_{t+1}$ produced by ForGAN (blue), G-regression (green) alongside the ground truth distribution (orange) for each time window cluster $c\,\in\,[y_{0_0},\ldots,y_{0_4}]$ and for the entire dataset on the Lorenz dataset.}
	\label{fig:reg_lorenz}
\end{figure*}

In the Lorenz experiment, the G-regression method performs better than GAN based on RMSE, MAE and MAPE values. However, we can perceive from Fig.~\ref{fig:reg_lorenz} how misleading these metrics can be. Fig.~\ref{fig:reg_lorenz} presents the probability distribution learned by ForGAN alongside the histogram of the G-regression predictions and the data distribution for each cluster on the test set as well as the entire test set. These plots indicate that ForGAN learns the probability distribution of the dataset precisely with respect to the corresponding cluster. Contrary, G-regression predictions are completely inaccurate While it obtained better scores on the point-wise metrics in comparison to ForGAN. The G-regression method has converged to the mean value of $x_{t+1}$ distribution for each cluster and as a result, the predictions do not represent the ground truth at all. Since the histogram of G-regression predictions does not cover the range of ground truth values, it is not possible to calculate KLD for G-regression method in this experiment.

\begin{figure*}
	\centering
	\includegraphics[width=\textwidth]{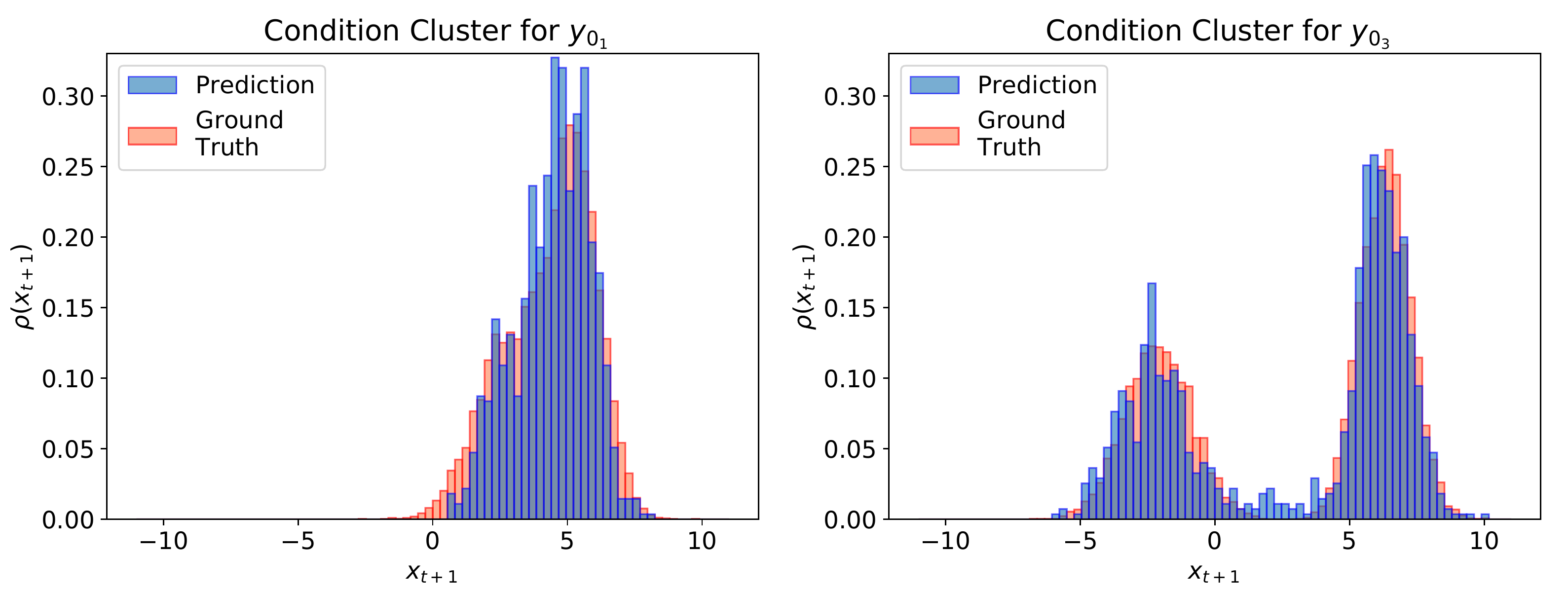}
	\caption{The probability distribution of $x_{t+1}$ learned by ForGAN for two randomly selected time windows $c$ and the data distribution of the time window cluster they origin from on Lorenz dataset.}
	\label{fig:lorenz_one_sample}
\end{figure*}

Furthermore, we expect ForGAN to forecast all possible outcomes for a given time window. To investigate the validity of this assumption, we select two random time windows from the test set and forecast $x_{t+1}$ 100 times using ForGAN. Fig.~\ref{fig:lorenz_one_sample} portrays the distribution of sampled $x_{t+1}$  alongside the probability distribution of their cluster. We can perceive from this figure that ForGAN can model the full probability distribution of $x_{t+1}$ for a given time window condition accurately.
\medskip

In the Mackey-Glass experiment, ForGAN outperforms both state-of-the-art~\cite{mendez2017competitive} and G-regression model based on point-wise error metrics as well as KLD. The G-regression model has the same structure as ForGAN and it is optimized directly on RMSE, yet ForGAN performs significantly better than G-regression. We find this observation to be the evidence for the effectiveness of adversarial training for forecasting in comparison to standard training methods.

\medskip

Finally, in our last experiment on Internet traffic dataset, G-regression method outperforms state-of-the-art and ForGAN based on the MAPE value. On the other hand, ForGAN performs almost two times better than G-regression method based on KLD. Furthermore, the KLD for the state-of-the-art method is not available. Due to inconsistency between point-wise error metrics and divergence measure, selecting the best method with certainty is not possible.
However, in the Lorenz experiment, we witness that it is possible to have a mean regression algorithm with a small point-wise error which is completely imprecise in forecasting future values. In any case, the performance of ForGAN on Internet traffic dataset is quite impressive. It outperforms the G-regression based on KLD and it falls behind other methods based on point-wise error metrics only with a narrow margin.
\section{Conclusion and Future Work}
\label{sec:conclusion}
We present ForGAN, a neural network for one step ahead probabilistic forecasting. Our method is trained using adversarial training to learn the conditional probability distribution of future values.

We test our method with three experiments. In the first experiment, ForGAN demonstrates its high capability of learning probability distributions while taking the input time window into account. In the next two experiments, ForGAN demonstrates impressive performance on two public datasets, showing the effectiveness of adversarial training for forecasting tasks.

We compare ForGAN to G-regression, where the generator architecture is kept, but RMSE loss is optimized. We demonstrate that while G-regression performs better than ForGAN based on some point-wise error metrics, it does not accurately model the real data distribution and ForGAN outperforms G-regression considering distribution divergence measure. Our experiments show that point-wise error metrics are not a precise indicator for the performance of forecasting methods. Furthermore, ForGAN demonstrates its high capability in forecasting full probability distribution of future values which makes it superior to conventional mean regression methods. Adversarial training enables us to train a model for probabilistic forecasting easily without facing any technical problems like quantile crossing nor any dependency on the chosen prior.

Our experiments reveal that in the presence of strong noise, the effectiveness of ForGAN is more prominent as we illustrate in Lorenz experiments. The performance of mean regression methods is close to ForGAN when the noise is weak. Since ForGAN can model data distributions with any level of noise, it is more reliable and a robust choice for forecasting in comparison to mean regression methods.

For future reference and comparison, we introduce the Lorenz dataset for probabilistic forecasting. It is based on a chaotic differential equation. The Lorenz dataset can be downloaded from \underline{https://cloud.dfki.de/owncloud/index.php/}\break\underline{s/KGJm5iNKrCnAwEg} .

\subsection{Future Work}
With the promising results from ForGAN, there are many possibilities to pursue this line of research further. One possible direction is to investigate the capability of ForGAN to forecast multiple-step ahead values. It would be interesting to find out how far we can push the horizon with ForGAN and compare to the state-of-the-art in multi-step prediction. Another direction is improving the architecture of ForGAN. In this paper, we limit the ForGAN structure to one layer of LSTM/GRU in generator and discriminator however one can study the performance of ForGAN with different architectures e.g. CNNs or different loss functions like Wasserstein~\cite{arjovsky2017wasserstein}. As in other works of studying GAN, we found some issues in evaluating and comparing the methods. Hence, further research in that direction would be beneficial, too.

\bibliographystyle{IEEEtran}
\bibliography{lite.bib}

\EOD

\end{document}